# Large Language Models as Tax Attorneys:
# A Case Study in Legal Capabilities Emergence

John J. Nay, David Karamardian, Sarah B. Lawsky, Wenting Tao, Meghana Bhat,
Raghav Jain, Aaron Travis Lee, Jonathan H. Choi, Jungo Kasai

Stanford University
University of Michigan
University of Washington
University of Southern California
Northwestern Pritzker School of Law
SimPPL

## Abstract

Better understanding of Large Language Models' (LLMs) legal analysis abilities can contribute to improving the efficiency of legal services, governing artificial intelligence, and leveraging LLMs to identify inconsistencies in law. This paper explores LLM capabilities in applying tax law. We choose this area of law because it has a structure that allows us to set up automated validation pipelines across thousands of examples, requires logical reasoning and maths skills, and enables us to test LLM capabilities in a manner relevant to real-world economic lives of citizens and companies. Our experiments demonstrate emerging legal understanding capabilities, with improved performance in each subsequent OpenAI model release. We experiment with retrieving and utilising the relevant legal authority to assess the impact of providing additional legal context to LLMs. Few-shot prompting, presenting examples of question-answer pairs, is also found to significantly enhance the performance of the most advanced model, GPT-4. The findings indicate that LLMs, particularly when combined with prompting enhancements and the correct legal texts, can perform at high levels of accuracy but not yet at expert tax lawyer levels. As LLMs continue to advance, their ability to reason about law autonomously could have significant implications for the legal profession and AI governance.

*Correspondence: John J. Nay, johnjnay.com, john.j.nay@gmail.com.

# Introduction

AI capabilities are marching forward.[1 2 3 4 5] Large Language Models (LLMs)[6] are the locus of the rapid advances. State-of-the-art LLMs can pass standardised tests[7] and plan, reason, and leverage tools.[8] LLMs, though, are essentially black boxes, even to their developers. We have little insight into their inner workings and have no guarantees on how an LLM will behave on a new task.[9 10 11] Best practice is to measure LLM performance on a litany of benchmarks before models are deployed beyond the research environment, but these benchmarks are often not real-world tasks we care about, or may have been memorised by the LLM during its training.[12] This phenomenon typically arises when the datasets used for training LLMs, often sourced from the internet, contain the same data used for performance evaluation. The overlap can inflate the estimate of the model's performance, giving an illusion of understanding that could instead be basic recognition.[13]

We focus evaluation effort specifically on legal analysis capabilities of LLMs for three reasons.

First, assessing the extent that LLMs grasp the law can contribute toward governing LLMs and automated systems more generally.[14] One policy-relevant approach seeks to leverage regulatory reasoning and legal reasoning within LLMs for "Law-Informed AI" aligned with societal values as determined by democratic processes and law-making. This "Law Informs Code" approach rests on the established effectiveness of the democratic process in creating adaptive legal standards such

---

[1] Benaich N, Hogarth I. State of AI Report 2022. STATEOF.AI [Internet]. 2022 Oct 11. Available from: https://www.stateof.ai/2022-report-launch.html

[2] Wei J. 137 Emergent Abilities of Large Language Models. JASONWEI.NET [Internet]. 2022 Nov 14. Available from: https://www.jasonwei.net/blog/emergence

[3] Webb T, et al. Emergent Analogical Reasoning in Large Language Models. ARXIV [Internet]. 2022 Dec 19. Available from: https://arxiv.org/abs/2212.09196

[4] Hafner D, et al. Mastering Diverse Domains through World Models. ARXIV [Internet]. 2023 Jan 10. Available from: https://arxiv.org/abs/2301.04104

[5] Institute for Human-Centered AI. STANFORD UNIV. [Internet]. [cited 2023 Mar 20]. Available from: https://hai.stanford.edu

[6] OpenAI. Better Language Models and Their Implications. OPENAI [Internet]. 2019 Feb 14. Available from: https://openai.com/research/better-language-models

[7] OpenAI. GPT-4. OPENAI [Internet]. 2023 Mar 14. Available from: https://openai.com/research/gpt-4

[8] Schick T, et al. Toolformer: Language Models Can Teach Themselves to Use Tools. ARXIV [Internet]. 2023 Feb 9. Available from: https://arxiv.org/abs/2302.04761

[9] World Economic Forum. The Promises and Perils of AI, with Prof Stuart Russell. WORLDECON. FORUM [Internet]. 2022 Jan 6. Available from: https://www.weforum.org/agenda/2022/01/artificial-intelligence-stuart-russell-radio-davos/

[10] Zhang D, et al. The AI Index 2022 Annual Report. STANFORD INST. FOR HUMAN-CENTERED AI, STANFORD UNIV. [Internet]. [cited 2023 Mar 20]. Available from: https://aiindex.stanford.edu/report/

[11] The New AI-Powered Bing is Threatening Users. That's no Laughing Matter. TIME [Internet]. 2023 Feb 17. Available from: time.com/6256529/bing-openai-chatgpt-danger-alignment/

[12] Magar, Inbal, and Roy Schwartz. "Data Contamination: From Memorization to Exploitation." *arXiv preprint arXiv:2203.08242* (2022).

[13] Chang, K. K., Cramer, M., Soni, S., & Bamman, D. (2023). Speak, memory: An archaeology of books known to chatgpt/gpt-4. arXiv preprint arXiv:2305.00118.

[14] Some examples in the US and UK include National Artificial Intelligence Initiative, About: National Artificial Intelligence Initiative, https://www.ai.gov/about/#naii-national-artificial-intelligence initiative. [Accessed 11 April 2023]; Advancing Trustworthy AI, National Artificial Intelligence Initiative, https://www.ai.gov/strategic pillars/advancing-trustworthy-ai. [Accessed 11 April 2023]; European Commission, Proposal for a Regulation of the European Parliament and of the Council Laying Down Harmonised Rules on Artificial Intelligence (Artificial Intelligence Act) and Amending Certain Union Legislative Acts, https://eur-lex.europa.eu/legal-content/EN/TXT/?uri=CELEX%3A52021PC0206. [Accessed 11 April 2023]

as fiduciary duties through iterative debate and litigation.[15] The premise is that learning the spirit of the law can guide AI systems in making reasonable choices in novel scenarios. For instance, LLMs exhibit an early ability to predict when fiduciary duties are violated,[16] and this capability could power safer AI deployments where an LLM-powered system serves a human principal.

Second, LLMs can be used as tools for humans to provide legal services more efficiently and effectively, whether that be self-service or through a professional attorney. If the models better understand law, they can be more reliable and ultimately more useful. LLMs can assist in tasks ranging from contract analysis to case prediction, potentially democratising access to legal advice, reducing the cost and complexity for those who might otherwise struggle to navigate the legal system. Rigorous safeguards should be put in place as these models are deployed, given the sensitive nature of legal work. This includes increasing data privacy, minimising bias, maintaining accountability for the decisions made with the help of these models, and evaluating the suitability of the LLMs for any given use case. Hence, the need for systematic evaluations.

Third, if LLMs understand the law well enough, they could be deployed by the government, citizens, and researchers to identify inconsistencies in existing laws.[17] LLMs could increase the efficiency and transparency of governments more broadly. For instance, LLMs can oftentimes provide clear, understandable explanations of complex laws and regulations. Eventually, LLMs may help predict likely impacts of new laws or policies. By scanning vast amounts of legal text and associated implementations, LLMs could flag potentially "outdated" law, or areas where the law is silent when, in other similar circumstances, the legislature or regulators provide guidance.

In this paper, we study retrieval-augmented generation of LLMs leveraging the text of the U.S. Code of Federal Regulations (CFR) and the U.S. Code (a compilation of federal statutes). We test the emerging capability of a suite of LLMs in understanding tax law.

We chose tax law for four reasons. First, unlike some legal subjects where the doctrines are distilled from numerous precedents, the legal authority in tax law is principally concentrated in two sources: the Treasury Regulations under the CFR and Title 26 of the U.S. Code (also called the Internal Revenue Code). This allows us to use a fixed universe of potentially relevant documents for the LLM's retrieval augmentation. Second, many tax rules allow for definitive answers to inquiries. This allows us to set up consistent and automated validation pipelines. Third, answering tax law questions for a given scenario usually requires logical reasoning skills and even maths skills beyond just reading the relevant legal authority, enabling us to test LLM capabilities in a manner relevant to real-world practice. Fourth, tax law is highly significant to the economic lives of nearly every citizen and company on a regular basis.

---

[15] Nay J. Law Informs Code: A Legal Informatics Approach to Aligning Artificial Intelligence with Humans. Northwestern J Tech & Intell Prop. 2023. https://scholarlycommons.law.northwestern.edu/njtip/vol20/iss3/1/

[16] Nay JJ. Large Language Models as Fiduciaries: A Case Study Toward Robustly Communicating with Artificial Intelligence through Legal Standards. SSRN [Internet]. 2023 Jan 30. Available from: https://papers.ssrn.com/sol3/papers.cfm?abstract_id=4335945

[17] Renée Jean, "Wyoming Lawmakers Turn To Experts To Stay Ahead Of AI Curve, Even As ChatGPT4 Passes Bar Exam" May 20, 2023 https://cowboystatedaily.com/2023/05/20/wyoming-lawmakers-turn-to-experts-to-stay-ahead-of-ai-curve-even-as-chatgpt4-passes-the-bar-exam/https://cowboystatedaily.com/2023/05/20/wyoming-lawmakers-turn-to-experts-to-stay-ahead-of-ai-curve-even-as-chatgpt4-passes-the-bar-exam/

We assess the accuracy of responses generated by LLMs on thousands of tax law inquiries across experimental setups: the use of the LLM alone and the integration of the LLM with the underlying legal texts, along with various retrieval techniques (with comparisons made across different retrieval methods). We conduct these experiments across LLMs, from smaller and weaker models, up through the largest state-of-the-art model, OpenAI's GPT-4. Each LLM we tested was state-of-the-art when it was originally released. Through examining results across increasingly large models, we find evidence for emerging legal understanding capabilities of LLMs, improving with each model release. This suggests that we may see the advent of superhuman AI legal skills as the state-of-the-art continues to rapidly advance.

## Our Approach to Validating LLM Legal Understanding

We test LLMs' ability to "understand" law. Giving relevant and correct legal advice for a specific situation is a task combining precise knowledge of legal sources as well as reasoning (and sometimes maths) capabilities to analyse situations.

We generate multiple-choice problems, each consisting of a question and a set of potential answers (only one of which is correct). The facts, names and numbers for each problem are randomly generated by Python code. As a result, our synthetic validation set consists solely of brand-new problems that do not exist on the internet and could not have been included in the training set for any LLM. This is an important distinction setting our validation apart from many other LLM benchmarking exercises. In many cases, the LLM being evaluated – which is trained on much of the internet – may have been trained on the validation data itself.

While some of the questions involve only qualitative understanding (e.g., "checking-the-box" questions, see Appendix for more), others involve more arithmetic operations as well (e.g., calculation to determine basis amount). The solution to a given problem can refer to either the CFR or the U.S. Code. To produce problems of a specific question type, we call our Python code to generate a bulk collection of multiple-choice problems. Each generated problem contains: (1) the legal question with answer options for the LLM to choose from; (2) the correct option; (3) the explanation for why that option is correct; and (4) the citation to the specific part of the law in which the answer to the question can be determined.

We generate two overarching multiple-choice exams for evaluation, one based on the CFR, and one based on the U.S. Code. The CFR and U.S. Code exams are composed of three and four sections, respectively, with each 100-question section pertaining to a specific tax law question type. See the Appendix for details on these seven types of questions across tax law categories.

For each question, we prompt an LLM to pick one of the multiple-choice answers, and we evaluate the LLM's performance based on whether it chooses the correct answer. Since manually grading over 28,000 questions across all experiments by hand is not feasible, and since the models don't always produce outputs in a consistent format that we can parse consistently and directly compare to the real answer, we utilise the most powerful available LLM, GPT-4, to compare the ground truth to the outputted answer. GPT-4 is instructed to grade the accuracy of a predicted answer choice by comparing it to the real answer choice for a given question.[18]

## Our Approach to Retrieval-Augmented Generation and LLM Prompting

We compare results across retrieval methods, each with its own prompt template that provides different supporting context to the LLM; see the Appendix for a full example of a prompt template from one of our experimental runs. When supplying supporting context to models with smaller context windows, we sometimes had to truncate the retrieved context to fit inside the window. The LLMs have the following context windows: davinci, 2049 tokens; text-davinci-002 and gpt-3.5-turbo, 4097 tokens; gpt-4, 8192 tokens.

Our first experimental setting for retrieval, "bypass_retrieval," creates a baseline for testing the impact of retrieval and LLM knowledge. In these cases, we simply provide the LLM with a multiple-choice question and the answer options with no additional explicitly provided legal context. This method assesses the ability of an LLM to answer a tax law question solely from its "internal knowledge base."

For the second retrieval experimental setting, "similarity_search," we inject potentially relevant legal text into the prompt. Offline, before running the evaluations, we extracted the statutes from Title 26 of the U.S. Code and regulations from the CFR Treasury Regulations in the form of discrete documents, each corresponding to a subsection from the legal source. The discrete subsection documents are roughly 130 tokens on average for our CFR vector database, and roughly 250 tokens on average for our U.S. Code vector database. We leverage the open-source "Facebook AI Similarity Search" library to create a vector database that maps the discrete subsections to 768-dimensional embeddings, computed by a state-of-the-art[19] dense retrieval model, GTR-large.[20] GTR-large is trained on large amounts of retrieval data from various domains, including biomedical and science text, but not legal text; thus, our retrieval use-case is "out-of-domain" for the embeddings model.[21] When a question is presented as input, our system retrieves the four most "relevant" documents from the vector store, where relevance is estimated based on the cosine similarity between the documents and the question. These documents are then injected as context into the prompt together with the original question, and the LLM is also instructed to return the metadata for which legal source subsections it referenced in its answer.

---

[18] Before deploying this evaluation method at scale to grade all of our experiments, we assessed its efficacy on a subset of questions across all the LLMs we were going to evaluate; our GPT-4 LLM evaluator perfectly graded nearly all of these questions, only erring less than 1% of times when grading the most ambiguous and difficult-to-grade outputs from the oldest and least capable model we evaluated, OpenAI's "davinci" release. For example, davinci can at times generate outputs with multiple conflicting answer choices (even when instructed not to), occasionally fooling our LLM grader that davinci correctly picked the one true answer. But this did not occur frequently enough to significantly impact the overall trends we report, which we determined by manually reviewing much of the davinci grading.

[19] Vaswani, A., Shazeer, N., Parmar, N., Uszkoreit, J., Jones, L., Gomez, A. N., Kaiser, L., & Polosukhin, I. (2017). Attention Is All You Need. In Advances in Neural Information Processing Systems.

[20] Ni, J., Qu, C., Lu, J., Dai, Z., Ábrego, G. H., Ma, J., Zhao, V. Y., Luan, Y., Hall, K. B., Chang, M.-W., Yang, Y. Large Dual Encoders Are Generalizable Retrievers. In: Proceedings of the 2022 Conference on Empirical Methods in Natural Language Processing.

The third experimental setting, "gold_truth," does not rely on a vector database or similarity search to provide the LLM with additional context; instead, we directly provide as context the correct source material, obtained by referencing each given question's citation to the specific part of law it pertains to. Incorporating this method in the experimental design helps estimate the impact of the theoretically best possible retrieval. Another way of looking at this design is that it isolates errors in the LLM's reasoning caused by inaccurate retrieval in the "similarity search" method.

For the fourth retrieval method, "lecture_notes," we provide context to the LLM in the form of lecture notes (written by Sarah Lawsky, one of this paper's co-authors and a Professor of Law at Northwestern) relevant to the given question type.

Another experimental variable was whether we employ "chain-of-thought" (CoT) prompting, which asks the LLM to think through its response step-by-step.

Finally, we experimented with few-shot prompting. This is where we provide a set of three other question-answer pair examples to the LLM, along with the question being asked. We varied the pairs to match the question type for the given problem and ensured that the question-answer pairs were not any of the questions used for evaluation. The notion behind few-shot prompting is to guide the LLM toward how to answer the given question by observing examples of how to answer questions. We did for all LLMs without providing contextual source documents or lecture notes.

## The LLMs

A primary factor we vary in our experiments is the LLM itself. We run the experimental design across four increasingly advanced LLMs released by OpenAI over the past three years. The weakest model we employ, 'davinci,' is the "most capable GPT-3 model." 'text-davinci-002' is an earlier version of GPT-3.5 that is "trained with supervised fine-tuning instead of reinforcement learning." 'gpt-3.5-turbo' is the "most capable GPT-3.5 model." The most capable model, 'gpt-4,' is "more capable than any GPT-3.5 model, able to do more complex tasks."[22]

For all models across all experiments, we set temperature equal to zero when generating responses to our prompts. Temperature is a parameter that controls the "randomness" of the model's output. For these LLMs, lower temperatures make the outputs more deterministic.

---

[21] We are developing "in-domain" embeddings currently for follow-up work.
[22] https://platform.openai.com/docs/models/

Finally, after running the experimental design across factors, we then run a final setting, "mega_run," which combines the "gold_truth" retrieval method, few-shot prompting, and CoT prompting of GPT-4. In other words, the most powerful combination of techniques and the most powerful model. This allows us to assess the upper bound on performance with these techniques. We discuss more advanced prompting in the Related Work and Next Steps sections that is likely to further boost performance.

Our total sample across the experiments contains 28,700 answers. Figure 1 below visualises the process and the experimental factors (in red).

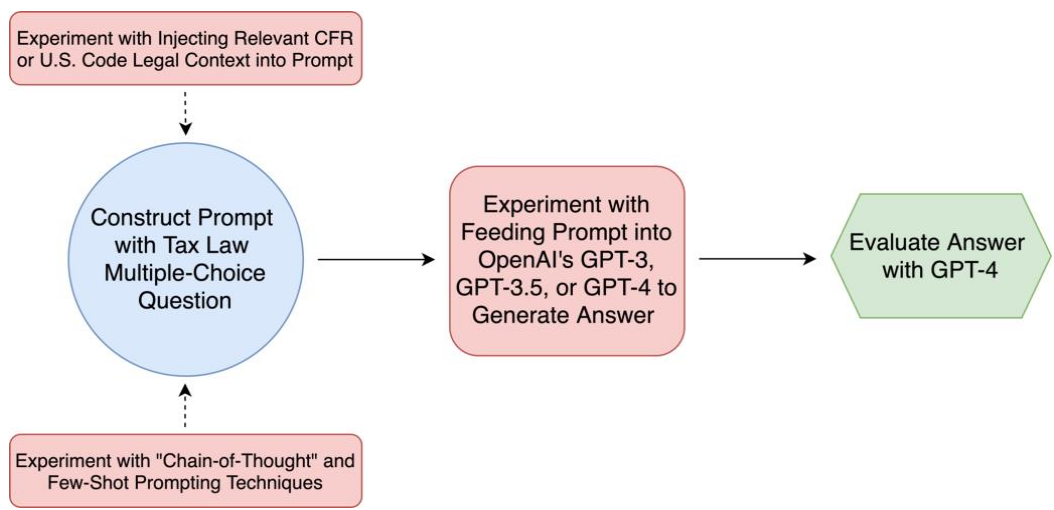

**Figure 1:** Our experimental pipeline compares performance on multiple-choice tax law exams across different LLMs, document retrieval techniques, and prompting techniques.

# Results

The first question we wanted to answer is whether CoT consistently improved the results for all (or most) models and all (or most) methods of retrieval. The answer is no, as evidenced by the difference between the solid and dashed lines in the charts of Figure 2. CoT does boost the performance of GPT-4, though. This suggests an LLM might need to have a certain capability level to be able to exhibit improved performance through additional reasoning. Two responses from GPT-4 prompted with CoT provide a sense of what our evaluation data looks like (Examples 1 and 2).

## Example 1:

*Question:* Aurora is hired by Grasshopper, Inc., on September 17, 2023. On her first day of work, when the stock trades at $18 per share, Aurora acquires 20 shares of Grasshopper stock for $15 per share, as contemplated by her employment agreement. The stock is a capital asset in Aurora's hands. As part of the terms of her employment, if she does not work for Grasshopper for 4 years, Aurora must return the stock. Anyone to whom Aurora transfers the stock is subject to the same restriction. Aurora makes no elections with respect to the stock. When the restriction expires and the stock vests in 2027, it is worth $23. Aurora sells the shares in 2028, when the stock is worth $25 per share. How much may Grasshopper deduct per share in 2027 due to the vesting of the shares?

(1) $0 (2) $2 (3) $3 (4) $7 (5) $8 (6) $10 (7) $15 (8) $18 (9) $23 (10) $25

*GPT-4 Response:* According to Section 83(a), the amount to be included in the gross income of the person who performed the services is the excess of the fair market value of the property at the time the rights become transferable or not subject to a substantial risk of forfeiture, over the amount paid for the property. In this case, the fair market value of the stock when it vests in 2027 is $23 per share, and Aurora paid $15 per share. The excess is $23 - $15 = $8 per share. Section 83(h) states that the employer, Grasshopper, can deduct an amount equal to the amount included in the gross income of the person who performed the services. In this case, that amount is $8 per share. So, Grasshopper may deduct (5) $8 per share in 2027 due to the vesting of the shares.

*Correct Answer:* (5) $8

## Example 2:

*Question:* Zahra owns property that is worth $41,000, with a basis of $33,000. Zahra dies and leaves the property to Dylan. Several years later, Dylan sells the property for $37,000. How much gain or loss does Dylan recognize due to his sale of the property for $37,000?

(1) -$4,000 (2) -$3,000 (3) -$488 (4) $0 (5) $2,000 (6) $4,000 (7) $14,000

*GPT-4 Response:* Zahra dies and leaves the property to Dylan. According to the context provided, the basis of the property in the hands of Dylan would be the fair market value of the property at the date of Zahra's death, which is $41,000. Now, Dylan sells the property for $37,000. To calculate the gain or loss, we subtract the basis from the selling price: $37,000 - $41,000 = -$4,000. So, the answer is (1) -$4,000.

*Correct Answer:* (1) -$4,000

The second experimental question is whether giving the LLM more legal text and more relevant (to the specific question we are asking it) legal text increases accuracy for all (or most) models. The answer is weakly yes, as evidenced by the upward slope of some of the lines in the charts of Figure 2. As you move from the left to the right in these charts, the experimental setup is providing more and/or more relevant legal text to the LLM, starting with the "bypass_retrieval" setup, where we don't give any source material to the LLM, to "gold_truth," where we give the most relevant source material to the LLM.

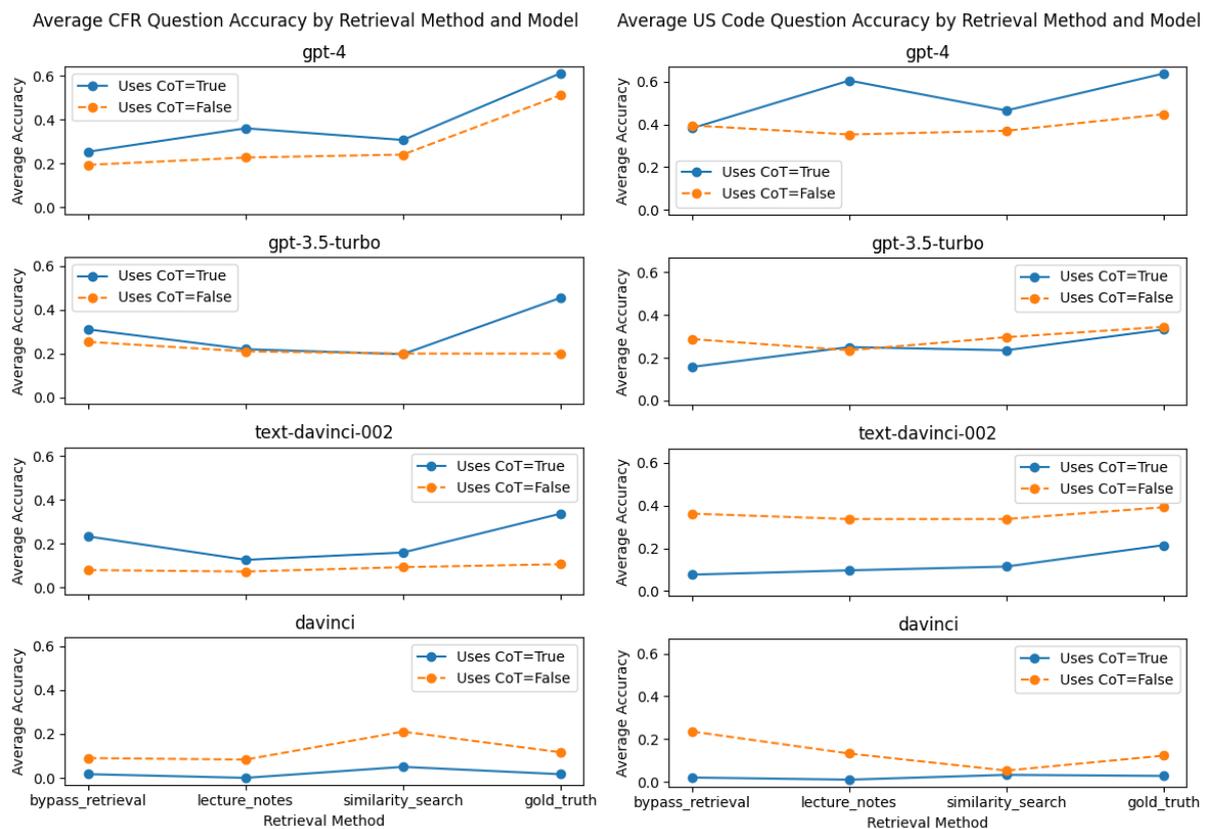

**Figure 2:** The y-axis is the accuracy of that experimental setting averaged across the different question sub-types. Spanning from left to right within each of the two columns, we generally

see a slight improvement in overall accuracy as the LLM is provided with more relevant legal source material with each subsequent retrieval method.

The third experimental question is whether few-shot prompting helps. The answer is a strong yes for GPT-4 and seems to be less consistently useful for weaker models. In the "few_shot" experimental setting, we don't give any source material to the LLM, but we input into the prompt examples of questions and answers from other questions that we are not testing it on, i.e., "few_shot" is "bypass_retrieval" plus few-shot prompting. The "mega_run" experiment combines "gold_truth" sources with few-shot and CoT prompting. As evidenced by Figure 3, GPT-4 can leverage relevant legal text and examples of the question-and-answer task to "reason" and come to a correct answer a large proportion of the time on difficult tax questions.

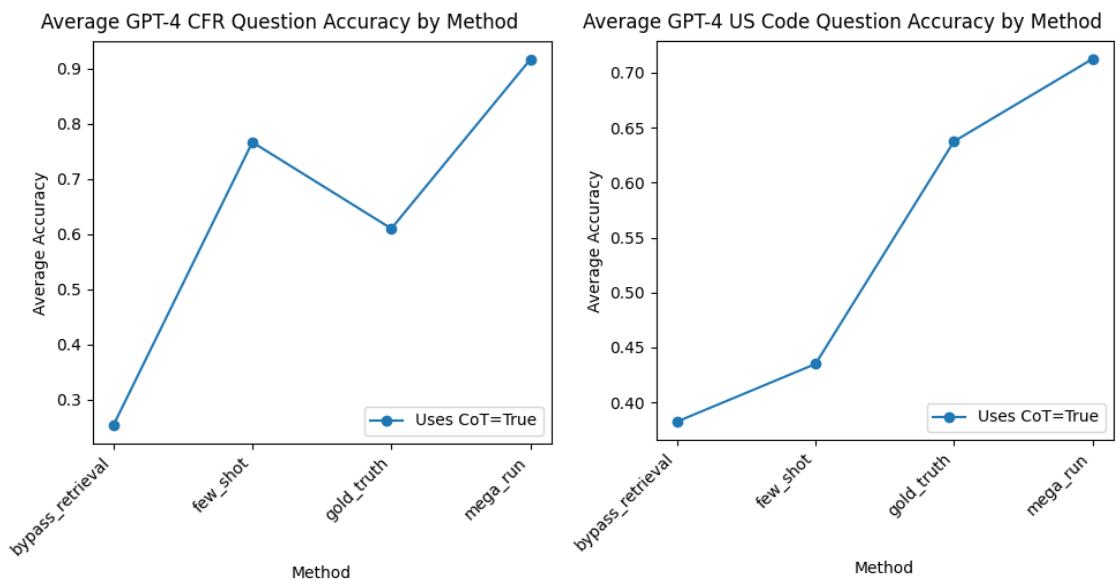

**Figure 3:** The y-axis is the accuracy of that experimental setting averaged across the different question sub-types. The "mega_run" experimental setup for GPT-4, which combines few-shot and CoT prompting, along with providing "gold truth" legal sources, results in best overall accuracy for both the CFR and U.S. Code exams. CoT boosts GPT-4 performance in the retrieval experimental settings of providing both no legal text ("bypass retrieval" and "few shot") and the most relevant possible legal text ("gold truth" and "mega run").

The primary experimental factor causing consistent increases in accuracy, when averaged across the other factors, is which underlying LLM is being used. This is consistent across the CFR and U.S. Code focused questions; the same pattern holds of newer models outperforming older models holds, as shown in Figure 4.

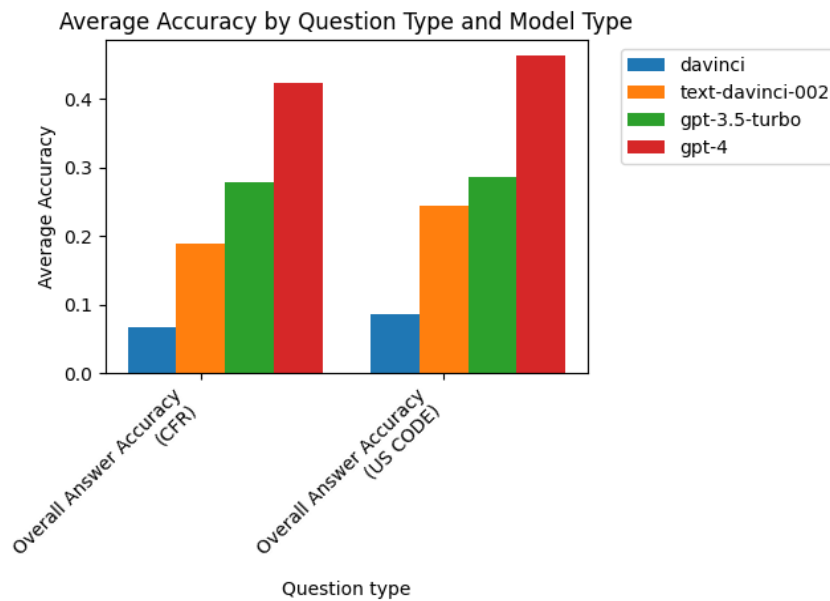

**Figure 4:** For both the CFR and U.S. Code exams, we see a clear increase in overall answer accuracy with each subsequently released OpenAI model. The most capable model, GPT-4, with both prompting enhancements (CoT and few-shot) and the most relevant "gold_truth" legal text input into the prompt, can perform extremely well, far better than any other setup in the experiments (see "mega_run" in Figure 3).

## Implications

Our work represents a step toward adapting LLMs to autonomously and reliably reason about law. While our experiments are limited to U.S. tax law, the capabilities the experiments investigate – finding relevant legal authorities and applying them to specific factual scenarios – are at the heart of legal work and could be generalised to other areas of legal practice. The increasing performance of LLMs on these tasks will have profound implications for the practice of law and the governance of AI.

First, lawyers are highly trained professionals, and LLMs could disrupt the legal services industry to the extent they are able to replicate much of the work of a skilled lawyer. We do not wish to overstate this possibility, since even our best current models underperform a professional tax lawyer, who would be expected to answer these questions with near-perfect accuracy. Moreover, answering clear-cut legal questions is only a small part of the work of a practising lawyer. Clients rely on lawyers for contextual advice, ethical counsel, and nuanced judgement, which at present LLMs are not able to provide as consistently as most human lawyers. Nevertheless, there is no strong reason to believe that LLMs could not eventually accomplish a wide range of legal tasks with greater performance, and our work represents a benchmark to track the improvement of LLMs at legal reasoning.

Second, even if LLMs are not *replacing* trained lawyers, they can *assist* a lawyer or provide a first draft of work a lawyer could subsequently check. This could significantly increase the productivity of practising lawyers and decrease the cost of legal services, potentially improving access to legal counsel for many people who currently cannot afford it. In addition, LLMs could provide useful legal information to consumers who are not engaging a traditional lawyer. As LLMs become more capable of autonomously providing basic legal advice, policymakers might have to reconsider regulations on how legal advice is delivered, including regulations about the unauthorised practice of law.

Third, governance is a key component of aligning AI with humans. Methods that improve LLM legal analysis skills are relevant, either by helping AI models "self-police" to ensure they are acting in accordance with law, or by designing separate models that can apply legal and ethical standards to confirm whether another AI is properly aligned with the law.

Our work also adds to the literature on emergent capabilities of LLMs by demonstrating the emergence of tax law understanding, which occurs once the LLM is of sufficient underlying general capability and is adequately prompted to elicit "reasoning" behaviour. Extrapolating these capabilities forward, LLMs being able to "understand" law would affect law-making [23] and necessitate changes to legal services regulation and emerging AI governance regimes.

## Related Work

LLM prompting involves designing text inputs to generate a response from an LLM. The goal of prompting is to steer the behaviour of the LLM in a way that elicits a desired outcome. Recent research has focused on developing effective prompting techniques that can expand LLMs' capabilities when carrying out a variety of tasks. Examples include prompt patterns,[24] in-context instruction learning,[25] evolutionary prompt engineering,[26] and domain-specific keywords with a trainable gated prompt to guide toward a target domain for general-domain LLMs.[27] Zhong et al.

---

[23] Nay JJ. Large Language Models as Corporate Lobbyists. [eprint]. arXiv. 2023 Jan 01 Available from: https://arxiv.org/abs/2301.01181.
[24] White J, Fu Q, Hays S, Sandborn M, Olea C, Gilbert H, et al. A Prompt Pattern Catalog to Enhance Prompt Engineering with ChatGPT [Internet]. arXiv [cs.SE]. 2023 [cited 2023 Mar 31]. Available from: https://arxiv.org/abs/2302.11382
[25] Ye S, Hwang H, Yang S, Yun H, Kim Y, Seo M. In-Context Instruction Learning [Internet]. arXiv [cs.CL]. 2023 [cited 2023 Mar 31]. Available from: https://arxiv.org/abs/2302.14691
[26] Chen A, Dohan DM, So DR. EvoPrompting: Language Models for Code-Level Neural Architecture Search [Internet]. arXiv [cs.NE]. 2023 [cited 2023 Mar 31]. Available from: https://arxiv.org/abs/2302.14838
[27] Goswami K, Lange L, Araki J, Adel H. SwitchPrompt: Learning Domain-Specific Gated Soft Prompts for Classification in Low-Resource Domains [Internet]. arXiv [cs.CL]. 2023 [cited 2023 Mar 31]. Available from: https://arxiv.org/abs/2302.06868

experiment with prompting LLMs to do scientific tasks across fields like business, science, and health by providing the LLM with a research goal and two large corpora, asking the LLM for corpus-level difference.[28] Reppert et. al develop iterated decomposition, a human-in-the-loop workflow for developing and refining compositional LLM programs that improves performance on real-world science question and answer tasks.[29]

More advanced techniques involve processes such as annotation, distillation, and model self-reflection. Diao et al. developed Active-prompt, which finds the most uncertain questions for the LLM and annotates those from the pool, achieving state-of-the-art on complex reasoning tasks.[30] Zhuo et al. develop methods for automatically designing multiple prompts and integrating automatic verbalizers without sacrificing performance.[31] LLMs can also improve through introspection. For example, Kim et al. develop a prompting scheme where an LLM agent recursively criticises and improves its output (RCI), outperforming supervised learning and reinforcement learning approaches on the MiniWoB++ benchmark, a web-based simulation task suite with tasks ranging from simple clicking to complex maths problems.[32] Press et al. investigate LLMs' ability to engage in compositional reasoning tasks, finding that as model size increases, single-hop question-answering ability improves more rapidly than multi-hop question-answering ability, resulting in a "compositionality gap." The authors propose "elicitive" prompting methods, such as CoT and self-talk, to mitigate this gap.[33] Yao et al. developed the popular ReAct approach where LLMs generate reasoning and actions in an interleaved manner, outperforming state-of-the-art baselines at the time across various tasks.[34] Jin et al. develop "Moral Chain-of-Thought" (MORALCoT) prompting, which draws from cognitive science theories of moral reasoning and excels in a novel challenge set centred on permissible rule-breaking.[35]

A growing body of research examines the characteristics of prompting. For instance, Lu et al. find that the performance of LLMs is not associated with the perceived difficulty of prompts estimated by human annotators, and that employing definitions, demonstrations, and explanations can enhance performance.[36] Halawi et al. investigate model performance when confronted with misleading or false prompts and reveal that LLMs exhibit comparable performance, irrespective of few-shot prompt accuracy, while accuracy discrepancies due to deceptive prompts only emerge in later layers of the model.[37] Focusing specifically on discrete prompts, Ishibashi et al. demonstrate that although these prompts exhibit a degree of robustness against certain perturbations, they remain vulnerable to others and fail to generalise effectively across natural language inference datasets. This underscores the necessity for further exploration into robust discrete prompting.[38] Focusing on the role of prompting in boosting LLMs' "Theory-of-Mind" performance, Moghaddam and Honey show that in-context learning prompts boost Theory-of-Mind abilities in GPT-4 and GPT-3.5 models.[39]

Prompting serves as a crucial element in utilising LLMs for real-world applications such as legal services and legal question-answering, as it connects model capabilities with targeted functionalities. In the context of our study, we examine LLMs' capacity to comprehend regulations and laws and experiment with the effects of very simple prompting techniques on accuracy. We leave the more advanced prompting discussed here for follow-up work in adapting these techniques to the legal domain.

Another burgeoning part of the LLM literature is dedicated to the capacity of LLMs to function as agents that perform tasks, make decisions, and interact with their environment. Andreas et al. demonstrate that LLMs can serve as agent models when only trained on bodies of documents, by implicitly inferring fine-grained communicative intentions and using that for subsequent text generation.[40] LLM-powered agents have demonstrated competence on some tasks that require reasoning, especially when combined with "tools" and symbolic systems. For instance, an AI system, Cicero, achieved human-level performance in the strategy game Diplomacy by integrating an LLM with strategic reasoning.[41] Furthermore, Shinn et al. explore LLM agents' ability for learning from mistakes with Reflexion, an approach that equips LLM-based agents with dynamic memory, "self-reflection" capabilities, and a method for detecting hallucinations.[42]

Regarding agentic LLMs more generally, Yang et al. investigate connections between LLMs and external entities, and their decision-making, using methods such as prompting, conditional generative modelling, planning, optimal control, and reinforcement learning.[43] A primary focus of autonomous agents lies in the interface between the LLM as an agent and the environment with which it interacts. Li et al. employ the "Internet Explorer" approach, which enables LLMs to dynamically use the internet as a continuously updating, open-ended dataset. In this approach, smaller models explore the web through self-supervision, locating relevant data to quickly enhance task performance.[44] Carta et al. examine a method to improve the alignment between the LLM's knowledge and its environment, while augmenting functional competence; the LLM is grounded in an interactive text world with online reinforcement learning, incrementally updating its knowledge based on observations.[45] Agents need to plan, and there is substantial interest in LLMs' ability to act as planners. Valmeekam et al. investigate the planning capabilities of LLMs, which exhibit poor performance in fully autonomous mode during common-sense planning tasks. However, when "heuristic guidance" and "human-in-the-loop" modes are employed, performance improves, albeit marginally.[46] As an example of a direction toward autonomous planning, Wang et al. developed a "Describe, Explain, Plan, and Select" approach, which explores the use of LLMs as planning agents in open-ended planning scenarios with long-term, multi-step tasks. This approach significantly improved performance in over 70 Minecraft tasks.[47] Other research examines LLMs as a component in building AI agents. For example, Li et al. explore the use of LLMs as probabilistic priors for generalised decision-making, applicable to non-linguistic perception and control, as well as tasks such as semantic segmentation, household navigation, and activity recognition.[48] The explosion of research interest at the intersection of autonomous agents and LLMs is relevant to our work, since agents that better understand the law are more likely to be aligned with society. By benchmarking legal understanding of LLMs, we can contribute to assessing the safety of agentic LLM deployments.

---

[37] Halawi D, Denain JS, Steinhardt J. Overthinking the Truth: Understanding how Language Models process False Demonstrations [Internet]. 2023 [cited 2023 Mar 31]. Available from: https://openreview.net/forum?id=em4xg1Gvxa

[38] Ishibashi Y, Bollegala D, Sudoh K, Nakamura S. Evaluating the Robustness of Discrete Prompts [Internet]. arXiv [cs.CL]. 2023 [cited 2023 Mar 31]. Available from: https://arxiv.org/abs/2302.05619

[39] Moghaddam SR, Honey CJ. Boosting Theory-of-Mind Performance in Large Language Models via Prompting. arXiv preprint arXiv:2304.11490. 2023.

[40] Andreas J. Language Models as Agent Models. arXiv preprint arXiv:2212.01681. 2022

[41] FAIR, Bakhtin A, Brown N, Dinan E, Farina G, Flaherty C, Fried D, Goff A, Gray J, Hu H, Jacob AP, Komeili M, Konath K, Kwon M, Lerer A, Lewis M, Miller AH, Mitts S, Renduchintala A, Roller S, Rowe D, Shi W, Spisak J, Wei A, Wu D, Zhang H, Zijlstra M. Human-level play in the game of Diplomacy by combining language models with strategic reasoning. Science. 2022 Nov 25;378(6624):1067-1074. Available from: https://www.science.org/doi/10.1126/science.ade9097. doi: 10.1126/science.ade9097. PMID: 34976589.

[42] Shinn N, Labash B, Gopinath A. Reflexion: an autonomous agent with dynamic memory and self-reflection. arXiv preprint arXiv:2303.11366. 2023 Mar 28

[43] Yang S, Nachum O, Du Y, Wei J, Abbeel P, Schuurmans D. Foundation Models for Decision Making: Problems, Methods, and Opportunities [Internet]. arXiv [cs.AI]. 2023 [cited 2023 Mar 31]. Available from: https://arxiv.org/abs/2303.04129

As LLMs demonstrate significant potential in tackling diverse tasks, research has focused on methods of evaluating their performance. Increasingly specific benchmarks are being developed. Examples include G-Eval, a framework using LLMs to evaluate natural language generation output via a CoT paradigm,[49] and AmbiEnt, where even advanced models like GPT-4 struggle with correctly disentangling ambiguous meanings.[50]

---

[44] Li AC, Brown E, Efros AA, Pathak D. Internet Explorer: Targeted Representation Learning on the Open Web [Internet]. arXiv [cs.LG]. 2023 [cited 2023 Mar 31]. Available from: https://arxiv.org/abs/2302.14051

[45] Carta T, Romac C, Wolf T, Lamprier S, Sigaud O, Oudeyer PY. Grounding Large Language Models in Interactive Environments with Online Reinforcement Learning. arXiv preprint arXiv:2302.02662. 2023 Feb

[46] Valmeekam K, Sreedharan S, Marquez M, Olmo A, Kambhampati S. On the Planning Abilities of Large Language Models (A Critical Investigation with a Proposed Benchmark) [Internet]. arXiv [cs.AI]. 2023 [cited 2023 Mar 31]. Available from: https://arxiv.org/abs/2302.06706

[47] Wang Z, Cai S, Liu A, Ma X, Liang Y. Describe, Explain, Plan and Select: Interactive Planning with Large Language Models Enables Open-World Multi-Task Agents. arXiv preprint arXiv:2302.01560. 2023 Feb 3

[48] Li BZ, Chen W, Sharma P, Andreas J. LaMPP: Language Models as Probabilistic Priors for Perception and Action. arXiv preprint arXiv:2302.02801. 2023 Feb 7.

[49] Liu Y, Iter D, Xu Y, Wang S, Xu R, Zhu C. G-Eval: NLG Evaluation using GPT-4 with Better Human Alignment. arXiv preprint arXiv:2303.16634. 2023

[50] Liu A, Wu Z, Michael J, Suhr A, West P, Koller A, Swayamdipta S, Smith NA, Choi Y. We're Afraid Language Models Aren't Modeling Ambiguity. arXiv preprint arXiv:2304.14399. 2023

Providing LLMs with domain-specific knowledge, updated data, and specialised reasoning and computation abilities can improve their performance on some tasks. Mialon et al. review the current advancements in augmentation, where LLMs are enhanced through reasoning capabilities, external modules, and tools. The authors argue that augmentation could potentially ameliorate interpretability, consistency, and scalability issues in LLMs.[51] Researchers have devised several methods for LLMs to employ external resources. For instance, Peng et al. introduce a system that employs plug-and-play external modules to refine grounded responses using external knowledge and iterative revision based on utility function feedback, substantially reducing LLM hallucinations.[52] Zhou et al. develop Doc-Prompting, a natural-language-to-code technique that utilises library documentation retrieval for code generation.[53] External documentation can also facilitate LLM self-assessment: Wu et al. establish a "Read and Reward" framework to enable an LLM to self-evaluate through manual learning. This framework employs a Question and Answer (QA) extraction module that condenses manual information and a reasoning module to assess interactions based on this information.[54]

QA has served as the testing ground for most of the LLM augmentation research thus far. Chen et al. review open-domain QA research.[55] Sil et al. introduced PRIMEQA, an open-source repository to democratise cutting-edge QA methodologies. This end-to-end QA toolkit allows for custom app creation with trainable retrievers and readers for deployment.[56] Sun et al. propose recitation-augmented language models, enabling LLMs to retrieve pertinent information from their own memory through sampling to answer questions.[57] Khattab et al. present Demonstrate-Search-Predict (DSP) for retrieval-augmented in-context learning that decomposes problems into more manageable components for both the language and retrieval models.[58] Ye et al. develop Compositional Exemplars for In-context Learning to assist in selecting the most diverse yet useful examples for LLMs to learn from for in-context learning.[59] Ram et al. present a simpler alternative to Retrieval-Augmented Language Modelling (RALM): in-context RALM, where grounding documents are incorporated into the LLM's input without modifying its architecture.[60] In our paper,

---

[51] Mialon G, Dessì R, Lomeli M, Nalmpantis C, Pasunuru R, Raileanu R, Rozière B, Schick T, Dwivedi-Yu J, Celikyilmaz A, Grave E, LeCun Y, Scialom T. Augmented Language Models: a Survey. arXiv. 2023 Feb 21. arXiv:2302.07842

[52] Peng B, Galley M, He P, Cheng H, Xie Y, Hu Y, et al. Check Your Facts and Try Again: Improving Large Language Models with External Knowledge and Automated Feedback. arXiv preprint. 2023 Feb 28:2302.12813

[53] Zhou S, Alon U, Xu FF, Wang Z, Jiang Z, Neubig G. DocPrompting: Generating Code by Retrieving the Docs. arXiv preprint arXiv:2207.05987. 2023

[54] Wu Y, Fan Y, Liang PP, Azaria A, Li Y, Mitchell TM. Read and Reap the Rewards: Learning to Play Atari with the Help of Instruction Manuals. arXiv preprint arXiv:2302.04449. 2023 Feb 1

[55] Chen D, Yih W. Open-Domain Question Answering. In: Proceedings of the 58th Annual Meeting of the Association for Computational Linguistics: Tutorial Abstracts. Online: Association for Computational Linguistics; 2020. p. 34-37. Available from: https://aclanthology.org/2020.acl-tutorials.8. doi: 10.18653/v1/2020.acl-tutorials.8

[56] Sil A, Sen J, Iyer B, Franz M, Fadnis K, Bornea M, et al. PrimeQA: The Prime Repository for State-of-the-Art Multilingual Question Answering Research and Development. arXiv preprint arXiv:2301.09715. 2023 Jan 26

[57] Sun Z, Wang X, Tay Y, Yang Y, Zhou D. Recitation-Augmented Language Models. arXiv preprint arXiv:2210.01296. 2023

[58] Khattab O, Santhanam K, Li XL, Hall D, Liang P, Potts C, Zaharia M. Demonstrate-Search-Predict: Composing retrieval and language models for knowledge-intensive NLP. arXiv preprint arXiv:2212.14024. 2023

[59] Ye J, Wu Z, Feng J, Yu T, Kong L. Compositional Exemplars for In-context Learning. arXiv preprint arXiv:2302.05698. 2023 Feb 15

[60] Ram O, Levine Y, Dalmedigos I, Muhlgay D, Shashua A, Leyton-Brown K, Shoham Y. In-Context Retrieval-Augmented Language Models. arXiv preprint arXiv:2302.00083. 2023 Feb

we focus on simple forms of augmentation, and leave testing these more sophisticated methods for future work.

There are many tasks that larger LLMs can complete that smaller models cannot.[61] Larger models have more inherent resources (for example, GPT-2 has 1.5 billion parameters while GPT-3 has 175 billion), and for some tasks that require various complex types of reasoning, LLMs' capability to do such tasks "emerges" in a nonlinear fashion after reaching a certain model size. Jason Wei has compiled a list of 137 emergent abilities of LLMs that have been uncovered by research, which includes things like "causal judgement" and "geometric shapes."[62] Our experiments suggest that legal understanding could be one such emergent ability.

## Next Steps

With clear evidence showing increases in capabilities from older to newer LLMs, attention can be shifted towards validating and improving the abilities of the newest, most powerful models available.

Regarding prompting, further analysis of our results could investigate the relationship between prompt length and accuracy. One possibility is that the LLMs do not perform as well as they could because their performance degrades as the length of the input increases; just because newer models like GPT-4 have a wider context window may not necessarily mean filling it to the max is optimal.

Many of the more advanced prompting techniques discussed in the Related Work section are prime candidates for increasing performance; in particular, the self-reflection and self-refinement techniques. For example, the LLM can be prompted with its own answers, and the relevant context, and asked, "Are there any ambiguities in this question that make it difficult to answer or for you to doubt your current answer? If so, conduct additional legal research by generating a topic that we need to search legal sources for." The response can then be used to conduct further retrieval augmented generation.

Regarding document retrieval, we seek to close the gap between the "similarity search" and "gold truth" retrieval methods through better retrieval. Especially for GPT-4, we saw a clear performance boost when feeding in the "gold truth" legal documents, rather than performing similarity search to extract the relevant documents from a vector database. This result indicates that our similarity search technique, and the various hyperparameter defaults we used, did not provide the most relevant "gold truth" sources into the LLM a significant portion of the time. Ultimately, as LLMs are deployed in real-world settings where humans won't be providing the exact legal documents necessary, the ability to retrieve the relevant documents will be important. We need to experiment with factors such as the choice of model embeddings, retrieval technique, and the token length of vector database subsections and number of subsections retrieved and placed into the prompt.

---

[61] Ornes S. The Unpredictable Abilities Emerging From Large AI Models [Internet]. Quanta Magazine. 2023 Mar 16 - [cited 2023 May 21]. Available from: https://www.quantamagazine.org/the-unpredictable-abilities-emerging-from-large-ai-models-20230316

[62] Wei J. 137 Emergent Abilities of Large Language Models [Internet]. Jason Wei Blog. [Date unknown] - [cited 2023 May 21]. Available from: https://www.jasonwei.net/blog/emergence

Finally, future work could compare performance between generally pre-trained LLMs, such as the OpenAI models in our experiments, and language models specifically pre-trained and fine-tuned for legal reasoning. Developing best practices for fine-tuning models for legal reasoning tasks is an important step towards sufficiently boosting AI legal capabilities in real-world settings.

## Acknowledgments

The Mercatus Center at George Mason University funded Meghana Bhat's work related to this research, and some of the computing costs for running the experiments. Pierre-Loic Doulcet, of Stanford University's CodeX Center for Legal Informatics, contributed to the U.S. Code vector embeddings work.

## Appendix

**CFR Exam Question Types:**
1. The first CFR question type, "Check the Box," involves entity status. The issue is whether a given business entity is eligible to select its tax status (e.g., partnership or disregarded entity), or whether it is required to be classified as a corporation. If it is eligible to select its tax status, the issue becomes what its default status would be and what status it would be able to elect into. Selecting a tax status is known colloquially as "checking the box."
2. The second CFR question type, "Restricted Property as Compensation/Employee," involves transfers of property as compensation. These problems ask how much of a deduction an employee is permitted to take when the employee receives property as compensation. The terms of the compensation require the employee to forfeit the property if certain conditions are not met.
3. The final CFR question type, "Part Gift/Part Sale," involves calculations related to the transfer of property that is in part a gift and in part a sale. These questions require determining how much gain a donor/seller recognizes on a transfer that is part gift/part sale, or the recipient's basis in the property that is received in a part gift/part sale transfer.

**U.S. Code Exam Question Types:**
1. The first U.S. Code question type, "Restricted Property as Compensation/Employer," is like the aforementioned CFR question type, "Restricted Property as Compensation/Employee," but instead pertains to how much of a deduction the employer is permitted to take due to transfer of restricted property as compensation.
2. The second U.S. Code question type, "Unrestricted Property as Compensation," asks about the treatment of employees who receive unrestricted property as compensation, in particular the amount of gain that these employees have when they sell the property they have received as compensation.

3.      The third U.S. Code question type is composed of three tax law areas: "Basis of Property Transferred as Gift," "Basis of Property Transferred at Death," and "Basis from Part Donation/Part Sale." These questions require determining the basis of property transferred as a gift, the basis of property that was acquired from a decedent, and how much gain a donor/seller recognizes on a transfer to a tax-exempt entity that is part donation/part sale.

4.      The final U.S. Code question type, "Qualified Business Income Deduction," requires determining the amount of deduction available under the qualified business income deduction provision, Section 199A.

## GPT-4 "mega_run" Prompt Template:

**System**: You are an AI assistant for answering questions about Tax Law.
Use the given pieces of context and multiple choice question-answer examples to answer the user's question.
If you don't know the answer, just say "I don't know the answer." Don't make up an answer.

**Human**: Use the provided context extracted from relevant documents and the following three multiple choice question-answer examples about Tax Laws to help you answer the fourth question at the end.

If you don't know the answer, just say "I don't know the answer." Don't make up an answer.
----------------
CONTEXT:
{insert "gold_truth" sources here}
----------------
QUESTION: De Dromedaris, vof, is organized in Netherlands as a Vennootschap onder firma. De Dromedaris has more than one member. At least one member of the entity has unlimited liability. Is De Dromedaris eligible to check the box? If so, what type of entity will it be if there is an election under the check-the-box rules (that is, if the entity does not take default status under the check-the-box rules)? (1) No, because a Vennootschap onder firma organized in Netherlands is a per se corporation. (2) Yes, and its elective status is a disregarded entity. (3) Yes, and its elective status is a partnership. (4) Yes, and its elective status is a corporation.
ANSWER: (4) Yes, and its elective status is a corporation.

QUESTION: Turkey, inc., is organized in Rhode Island as a corporation. Turkey has more than one member. All members of Turkey have limited liability. Is Turkey eligible to check the box? If so, what is its default status if there is no election under the check-the-box rules? (1) No, because a corporation organized in Rhode Island is a per se corporation. (2) Yes, and its default status is a disregarded entity. (3) Yes, and its default status is a partnership. (4) Yes, and its default status is a corporation.
ANSWER: (1) No, because a corporation organized in Rhode Island is a per se corporation.

QUESTION: El Pollo, sa, is organized in Colombia as a Sociedad Anonima. El Pollo has more than one member. All members of El Pollo have limited liability. Is El Pollo eligible to check the box? If so, what is its default status if there is no election under the check-the-box rules? (1) No, because a Sociedad Anonima organized in Colombia is a per se corporation. (2) Yes, and its default status is a disregarded entity. (3) Yes, and its default status is a partnership. (4) Yes, and its default status is a corporation.
ANSWER: (1) No, because a Sociedad Anonima organized in Colombia is a per se corporation.
----------------
QUESTION: {insert actual question here}
ANSWER:
Let's think step by step.